\documentclass[letterpaper]{article}
\usepackage{aaai18}
\usepackage{times}
\usepackage{helvet}
\usepackage{courier}
\usepackage{url}
\usepackage{graphicx}

\usepackage{soul}
\usepackage{color}
\usepackage{amsmath}
\usepackage{amssymb}
\usepackage{amsthm}
\usepackage{amsfonts}
\usepackage{array}

\title{How Images Inspire Poems: Generating Classical Chinese Poetry from Images with Memory Networks}
\author{Linli Xu$^{\dag}$, Liang Jiang$^{\dag}$, Chuan Qin$^{\dag}$, Zhe Wang$^{\ddag}$, Dongfang Du$^{\dag}$\\
$^\dag$Anhui Province Key Laboratory of Big Data Analysis and Application,\\
School of Computer Science and Technology, University of Science and Technology of China\\
$^\ddag$AI Department, Ant Financial Services Group\\
linlixu@ustc.edu.cn, jal@mail.ustc.edu.cn, chuanqin0426@gmail.com\\
wz143459@antfin.com, dfdu@mail.ustc.edu.cn
}
\begin{document}
\maketitle
\begin{abstract}
With the recent advances of neural models and natural language processing, automatic generation of classical Chinese poetry has drawn significant attention due to its artistic and cultural value. Previous works mainly focus on generating poetry given keywords or other text information, while visual inspirations for poetry have been rarely explored. Generating poetry from images is much more challenging than generating poetry from text, since images contain very rich visual information which cannot be described completely using several keywords, and a good poem should convey the image accurately. In this paper, we propose a memory based neural model which exploits images to generate poems. Specifically, an Encoder-Decoder model with a topic memory network is proposed to generate classical Chinese poetry from images. To the best of our knowledge, this is the first work attempting to generate classical Chinese poetry from images with neural networks. A comprehensive experimental investigation with both human evaluation and quantitative analysis demonstrates that the proposed model can generate poems which convey images accurately.
\end{abstract}

\section{Introduction}
Classical Chinese poetry is a priceless and important heritage in Chinese culture. During the history of more than 2000 years in China, millions of classical Chinese poems have been written to praise heroic characters, beautiful scenery, love, etc. Classical Chinese poetry is still fascinating us today with its concise structure, rhythmic beauty and rich emotions. There are distinct genres of classical Chinese poetry, including Tang poetry, Song iambics and Qing poetry, etc., each of which has different structures and rules. Among them, \textit{quatrain} is the most popular one, which consists of four lines, with five or seven characters in each line. The lines of a quatrain follow specific rules including the regulated rhythmical pattern, where the last characters in the first (optional), second and fourth line must belong to the same rhythm category. In addition, each Chinese character is associated with one tone which is either \textit{Ping} (the level tone) or \textit{Ze} (the downward tone), and quatrains are required to follow a pre-defined tonal pattern which regulates the tones of characters at various positions~\cite{wang2002summary}. An example of a quatrain written by a very famous classical Chinese poet Li Bai is shown in Table~\ref{quatrain}.

\begin{table}
\centering
\begin{tabular}{|c|} \hline
 \\
\includegraphics[scale = 2.2]{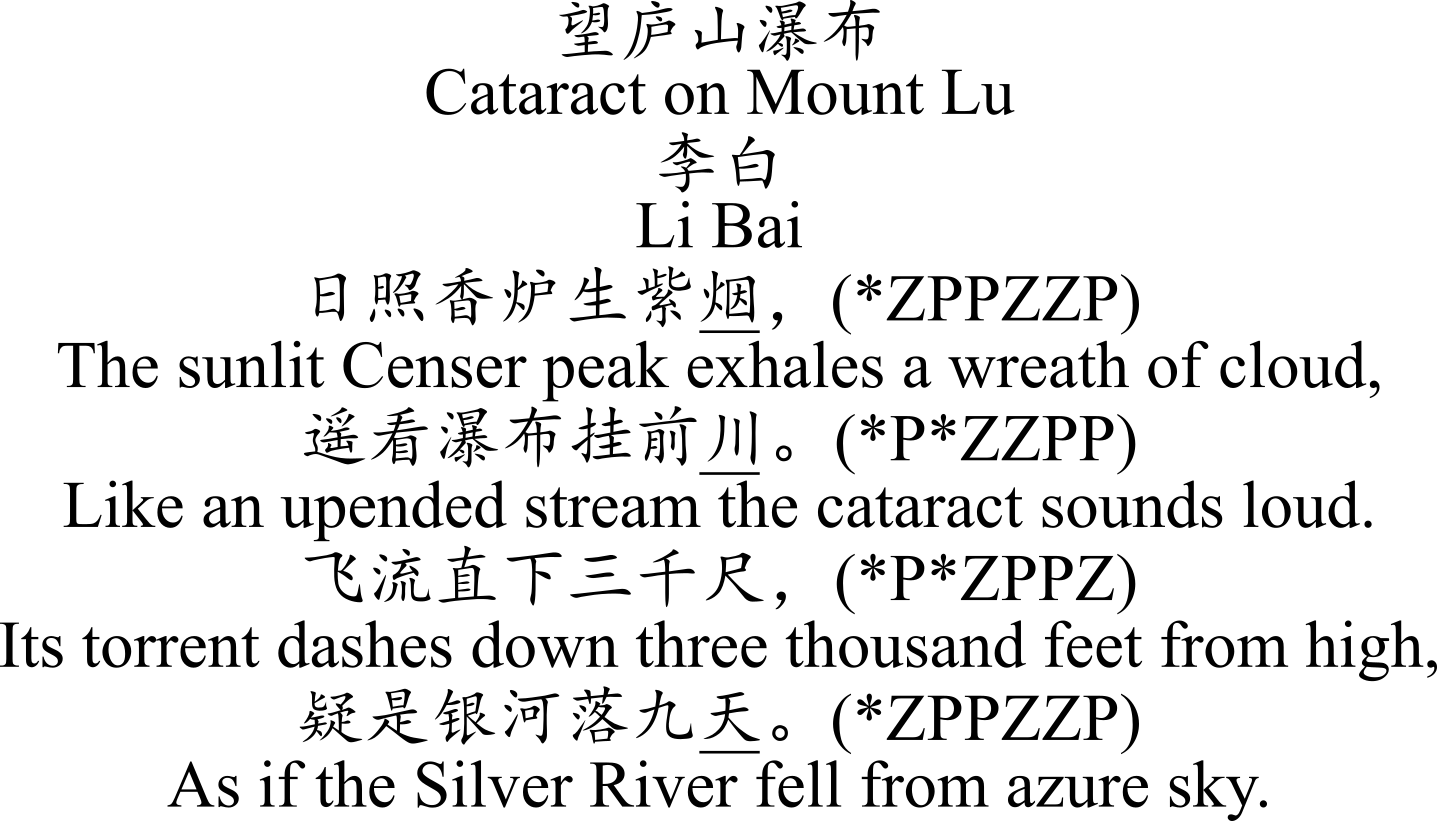} \\
\hline
\end{tabular}
\caption{An example of a 7-char quatrain. The tone of each character is shown at the end of each line, where ``P" represents \textit{Ping} (the level tone), ``Z" represents \textit{Ze} (the downward tone), and ``*" indicates the tone can be either. Rhyming characters are underlined.}
\label{quatrain}
\end{table}

The stringent rhyme and tone regulations of classical Chinese poetry pose a major challenge for generating Chinese poems automatically. In recent years, various attempts have been made on automatic generation of classical Chinese poems from text 
information 
such as keywords. Among them, rule-based approaches~\cite{tosa2008hitch,wu2009new,netzer2009gaiku,oliveira2009automatic,oliveira2012poetryme}, genetic algorithms~\cite{manurung2004evolutionary,zhou2010genetic,manurung2012using} and statistical machine translation methods~\cite{jiang2008generating,he2012generating} have been developed. More recently, with the significant advances in deep neural networks, a number of poetry generation algorithms based on neural networks have been proposed with the paradigm of sequence-to-sequence learning, where poems are generated line by line and each line is generated by taking the previous lines as input~\cite{zhang2014chinese,yi2016generating,wang2016chinese,zhewang2016chinese,zhang2017flexible}. However, restrictions exist in previous works, including topic drift and semantic inconsistency which are caused by only considering the writing intent of a user in the first line. In addition, a limited number of keywords with a proper order is usually required~\cite{zhewang2016chinese}, which limits the flexibility of the process.

On the other hand, visual inspirations are more natural and intuitive than text for poem writing. People can write poems to express their aesthetic appreciation or sentimental reflections at the sight of absorbing scenery, such as magnificent mountains and fast rivers. As a consequence, there usually exists a correspondence between a poem and an image explicitly or implicitly, where a poem either describes a scene, or leaves visual impressions on the readers. For example, the poem shown in Table~\ref{quatrain} represents an image with a magnificent waterfall falling down from a very high mountain. Because of this inherent correlation, generating classical Chinese poems from images becomes an interesting research topic.

As far as we know, visual inspirations from images have been rarely explored in automatic generation of classical Chinese poetry. The task of generating poetry from images is much more challenging than generating poetry from keywords in general, since very rich visual information is contained in an image, which requires sophisticated representation as a bridge to convey the essential visual features and semantic concepts of the image to the poem generator. In addition, to generate a poem that is consistent with the image and coherent itself, the topic flow should be carefully manipulated in the generated sequence of characters.

In this paper, we propose an Encoder-Decoder framework with topic memory to generate classical Chinese poetry from images, which integrates direct visual information with semantic topic information in the form of keywords extracted from images. We consider poetry generation as a sequence-to-sequence learning problem, where a poem is generated line by line, and each line is generated by taking all previous lines into account. Moreover, we introduce a memory network to support unlimited number of keywords extracted from images and to determine a latent topic for each character in the generated poem dynamically.
This addresses the issues of topic drift and semantic inconsistency, while resolving the restriction of requiring a limited number of keywords with a proper order in previous works. Meanwhile, to leverage the visual information of images not contained in the semantic concepts, we integrate direct visual information into the proposed Encoder-Decoder framework to ensure the correspondence between images and poems. The experimental results demonstrate that our model has great advantages in exploiting the visual and topic information in images, and it can generate poetry with high quality which conveys images accurately and consistently.

\noindent The main contributions of this paper are:
\begin{enumerate}
\item We consider a new research topic of generating classical Chinese poetry from images, which is not only for entertainment or education, but also an exploration integrating natural language processing and computer vision.
\item We employ memory networks to tackle the problems of topic drift and semantic inconsistency, while resolving the restriction of requiring a limited number of keywords with a proper order in previous works.
\item We integrate keywords summarizing semantic topics and direct visual information to capture the information conveyed in images when generating poems.
\end{enumerate}
\begin{figure*}[t]
\centering
\includegraphics[scale = 0.35]{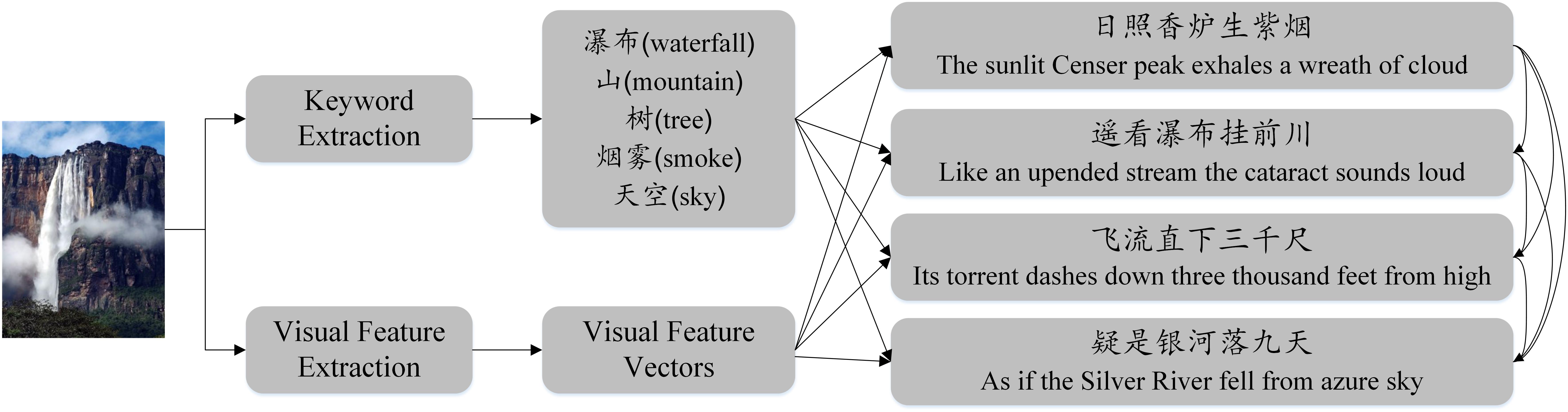}
\caption{An illustration of the pipeline to generate a poem from an image using Memory-Based Image to Poem Generator.}
\label{framework}
\end{figure*}

\section{Related Work}
Poetry generation has been a challenging task over the past decades. A variety of approaches have been proposed, 
most of which focus on generating poetry from text. Among them,
phrase search based approaches are proposed in \cite{tosa2008hitch,wu2009new} for Japanese poetry generation. Semantic and grammar templates are used in~\cite{oliveira2009automatic}, while genetic algorithms are employed in~
\cite{manurung2004evolutionary,manurung2012using,zhou2010genetic}. In the work of ~\cite{jiang2008generating,zhou2010genetic,he2012generating}, poetry generation is treated as a statistical machine translation problem, where the next line is generated by translating the previous line.
 Another approach in~\cite{yan2013poet} generates poetry by summarizing users' queries.

More recently, deep neural networks have been applied in automatic poetry generation. An RNN-based framework is proposed in~\cite{zhang2014chinese} 
where each poem line is generated by taking the previously generated lines as input. In the work of~\cite{yi2016generating}, the attention mechanism is introduced into poetry generation, where an attention-based Encoder-Decoder model is proposed to generate poem lines sequentially. A different genre of classical Chinese poetry is generated in \cite{wang2016chinese}, which is the first work to generate Chinese Song iambics, with each line of variable length. However, the neural methods introduced above share the limitation that the writing intent of a user is taken into consideration only in the first line, which will cause topic drift in the following lines. To address that, a modified attention-based Encoder-Decoder model is presented in \cite{zhewang2016chinese} which assigns a sub-topic keyword for each poem line. As a consequence, the number of keywords is fixed to the number of poem lines, and the keywords have to be sorted in a proper order manually, which limits the flexibility of the approach.

In this work, we address the issues in the previous works by introducing a memory network with unlimited capacity of keywords which can dynamically determine a topic for each character.
Memory networks are a class of neural networks proposed in~\cite{weston2014memory} to augment
recurrent neural networks with a memory component to model long term memory. In the work of
~\cite{sukhbaatar2015end}, memory networks are extended with an end-to-end training mechanism, which makes the model more generally applicable.
\citeauthor{zhang2017flexible}~(\citeyear{zhang2017flexible}) introduces memory networks into automatic poetry generation for the first time to leverage the prior knowledges in the existing poems while writing new poems. In the testing stage, by using a memory network, some existing poems are represented as external memory which provides prior knowledge 
for new poems. In contrast to \cite{zhang2017flexible}, we employ
memory networks in a completely different way, where we represent keywords as memory entities, such that we can handle keywords without restrictions and dynamically determine the topic for each character.

\section{Memory-Based Image to Poem Generator}
Images convey rich information from visual signals to semantic topics that can inspire good poems. To build a natural correspondence between an image and a poem, we propose a framework which integrates the semantic keywords summarizing the important topics of an image, along with the direct visual information to generate a poem. Specifically, given an image, keywords containing semantic topics are extracted as the outline when generating the poem, while visual information is exploited to embody the information not conveyed by the keywords.

\subsection{Framework}

Given an image $I$, we are generating a poem $P$ which consists of $L$ poem lines $\{l_1,l_2,..,l_L\}$. In the framework as illustrated in Figure~\ref{framework},
we first extract a set of keywords $K=\{k_1,k_2,...,k_N\}$ and a set of visual feature vectors $V=\{v_1,v_2,...,v_B\}$ from $I$ with keyword extractor and a convolutional neural network (CNN) based visual feature extractor respectively. The poem is then generated line by line. Specifically, when generating the $i$-th line $l_i$, the previously generated lines denoted by $l_{1:i-1}$, which is the concatenation from $l_1$ to $l_{i-1}$, the keywords $K$ and visual feature vectors $V$ are jointly exploited in
the Memory-based Image to Poem Generator (MIPG) model, which is the key component in the framework.

\begin{figure*}[t]
\centering
\includegraphics[scale = 0.6]{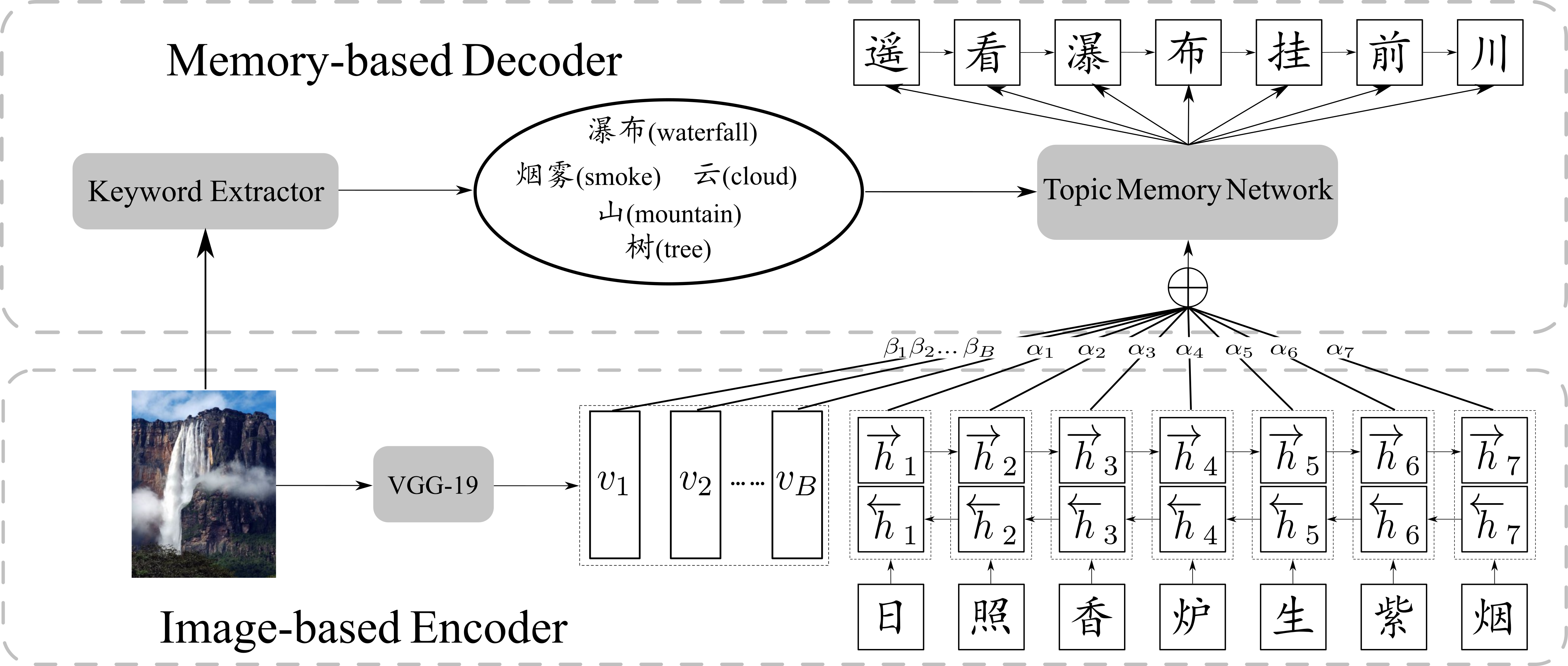}
\caption{An illustration of the Memory-based Image to Poem Generator (MIPG).}
\label{model}
\end{figure*}

As illustrated in Figure~\ref{model}, the MIPG model is essentially an Encoder-Decoder model consisting of two modules: an Image-based Encoder (I-Enc) and a Memory-based Decoder (M-Dec). In I-Enc, visual features $V$ are extracted from 
the image $I$ with a CNN, while a bi-directional Gated Recurrent Unit (Bi-GRU) model~\cite{cho2014learning} is employed to build semantic features $H$ from previously generated lines $l_{1:i-1}$. In M-Dec, the $i$-th poem line, denoted by a sequence of characters $\{y_1,...,y_G\}$, is generated based on the keywords $K$, the visual feature vectors $V$, as well as the semantic features $H$ from the previous lines. To generate
each character $y_t\in l_i$, we first transform $V$ and $H$ into dynamic representations for $y_t$ with the attention mechanism, followed by a Topic Memory Network designed to dynamically determine a topic for $y_t$. Finally $y_t$ is predicted using a topic-bias probability which enhances the consistency between the image and poem.

\subsection{Image-based Encoder (I-Enc)}
In I-Enc, as illustrated in the lower half of Figure~\ref{model}, we first encode the image $I$ into $B$ local visual features vectors $V=\{v_1,v_2,...,v_{B}\}$, each of which is a $D_v$-dimensional representation corresponding to a different part of the image. We use a CNN model as the visual feature extractor. 
Specifically, $V$ is produced by fetching the output of a certain convolutional layer of the CNN feature extractor which takes $I$ as input£º
$$
V=\mathrm{CNN}(I).
$$
Meanwhile, we use a Bi-GRU to encode the preceding lines of the generated poem $l_{1:i-1}$, which takes the form of a sequence of the character embeddings
$\{x_1,x_2,...,x_C\}$ into the corresponding hidden vectors $H=[h_1,h_2,...,h_C]$, where $C$ denotes the length of $l_{1:i-1}$ and $h_j$ is the concatenation of the forward hidden vector $\overrightarrow{h}_j$ and backward hidden vector $\overleftarrow{h}_j$ at the $j$-th step in the Bi-GRU. That is,
\begin{equation*}
\begin{gathered}
\overrightarrow{h}_j=\mathrm{GRU}(\overrightarrow{h}_{j-1},x_j), \\
\overleftarrow{h}_j=\mathrm{GRU}(\overleftarrow{h}_{j+1},x_j), \\
h_j=[\overrightarrow{h}_j;\overleftarrow{h}_j ].
\end{gathered}
\end{equation*}
The visual features $V$ and semantic features $H$ extracted in I-Enc are then exploited in M-Dec to dynamically determine which parts of the image and what content in the preceding lines it should focus on when generating each character.

\subsection{Memory-based Decoder (M-Dec)}
In M-Dec, as illustrated in the upper half of Figure~\ref{model}, each line $l_i=\{y_1,...,y_G\}$ is generated character by character. Specifically, at the $t$-th step, we use another GRU which maintains an internal state $s_t$ to predict $y_t$. $s_t$ is updated 
based on $s_{t-1}$, $y_{t-1}$, $H$ and $V$ recurrently, and can be formulated as
\begin{equation*}\label{decoder_basic}
s_t=f(s_{t-1},y_{t-1},\hat{h}_t,\hat{v}_t),
\end{equation*}
\begin{equation*}
\hat{h}_t=\mathrm{att}_H(H),
\end{equation*}
\begin{equation*}
\hat{v}_t=\mathrm{att}_V(V),
\end{equation*}
where $f$ denotes the function to update the internal state in GRU, $\mathrm{att}_H$ and $\mathrm{att}_V$ denote the functions of attention which transform $H$ and $V$ into dynamic representations $\hat{h}_t$ and $\hat{v}_t$ respectively, indicating which parts of the image and what content in the preceding lines the model should focus on when generating the next character. More formally,
%
\begin{equation*}\label{attention_h}
\hat{h}_t=\mathrm{att}_H(H)=\sum\limits_{j=1}^{C}\alpha_{tj}h_j,\quad h_j\in H,
\end{equation*}
where $\alpha_{tj}$ is the weight of the $j$-th character computed by the attention model:
\begin{equation*}
\alpha_{tj}=\frac{\exp(a_{tj})}{\sum\limits_{n=1}^{C}\exp(a_{tn})},
\end{equation*}
\begin{equation*}
a_{tn}=u_a^T\tanh(W_as_{t-1}+U_ah_n),
\end{equation*}
in which $a_{tj}$ is the attention score on $h_j$ at the $t$-th step. $u_a$, $W_a$ and $U_a$ are the parameters to learn.

Similarly, $\hat{v}_t$ is computed using $\mathrm{att}_V$ taking $V$ as input:
\begin{equation*}
\label{attention_v}
\hat{v}_t=\mathrm{att}_V(V)=\sum\limits_{j=1}^{B}\beta_{tj}v_j, \quad v_j\in V,
\end{equation*}
\begin{equation*}
\beta_{tj}=\frac{\exp(b_{tj})}{\sum\limits_{n=1}^B\exp(b_{tn})},
\end{equation*}
\begin{equation*}
b_{tn}=u_b^T\tanh(W_bs_{t-1}+U_bv_n).
\end{equation*}
Instead of predicting the $t$-th character $y_t$ by using $s_t$ directly, we introduce a \textbf{Topic Memory Network} to dynamically determine a proper topic for $y_t$ by taking $s_t$ as input and output a topic-aware state vector $o_t$ which contains not only information in the image and the preceding lines, but also the latent topic to generate $y_t$. A multi-layer perceptron is then used to predict $y_t$ from $o_t$.

\subsubsection{Topic Memory Network.}
\begin{figure}[t]
\centering
\includegraphics[scale = 0.7]{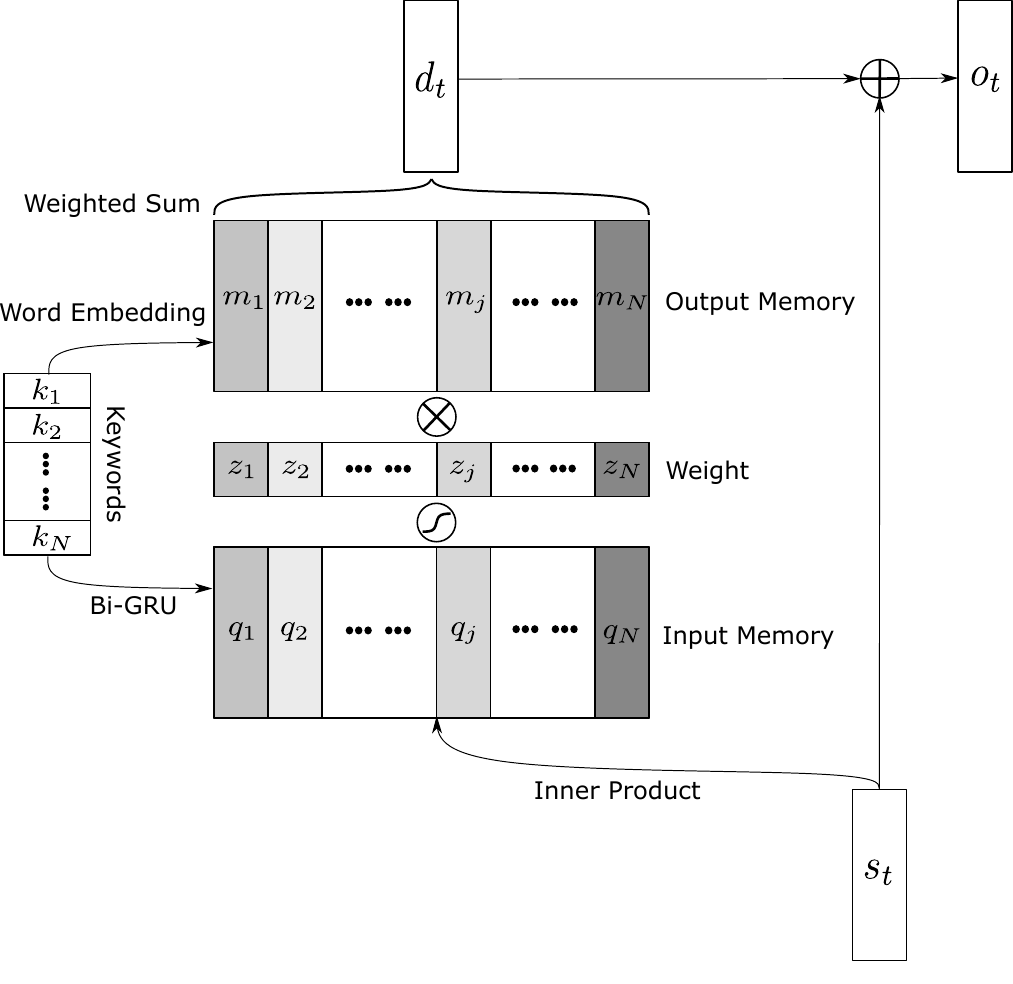}
\caption{An illustration of the Topic Memory Network.}
\label{memnn}
\end{figure}

Considering the rich information conveyed by images, it is essentially impossible to fully describe an image with too few keywords. In the meantime, the problem of topic drift would weaken the consistency between a pair of image and poem. To address these issues, as illustrated in Figure \ref{memnn}, we use a Topic Memory Network, in which each memory entity is a keyword extracted from an image, to dynamically determine a latent topic for each character by taking all keywords extracted from the image into consideration.




To achieve that, we use the \textit{general-v1.3 model} provided by Clarifai\footnote{https://clarifai.com} to extract a set of keywords $K=\{k_1,...,k_N\}$, and encode each keyword $k_j$ in $K$ into two memory vectors: an input memory vector $q_j$ which is used to calculate the importance of $k_j$ on $y_t$, and an output memory vector $m_j$ which contains the semantic information of $k_j$.

Specifically, for each keyword $k_j\in K$ with $C_j$ characters, we encode $k_j$ into a semantic vector. To take the order of $C_j$ characters into account, we use a Bi-GRU to encode $k_j$ into a sequence of forward hidden state vectors $[\overrightarrow{q_1},...,\overrightarrow{q_{C_j}}]$ and a sequence of backward hidden state vectors $[\overleftarrow{q_1},...,\overleftarrow{q_{C_j}}]$. Then the input memory vector $q_j$ is computed by concatenating the last forward state $\overrightarrow{q_{C_j}}$ and the first backward state $\overleftarrow{q_1}$, that is, $q_j=[\overrightarrow{q_{C_j}};\overleftarrow{q_1}]$.

The output memory representation $m_j$ of each keyword $k_j$ is computed by the mean value of the embedding vectors of all characters in $k_j$,
$$
m_j=\frac{1}{C_j}\sum\limits_{n=1}^{C_j}e_{jn},
$$
where $e_{jn}$ represents the word embedding of the $n$-th character in $k_j$ that is learned during training.

With the hidden state vector at the $t$-th step $s_t$, we compute the importance $z_j$ of each keyword $k_j\in K$ based on the similarity between $s_t$ and the input memory representation $q_j$ of 
the keyword. This can be formulated as follows:
$$
z_j=\mathrm{softmax}(s_t^Tq_j),\quad 1\leq j\leq N.
$$
Given the weights of the keywords $[z_1,...,z_N]$, a latent topic vector $d_t$ for $y_t$ is calculated by the weighted sum of the output memory representations of keywords $[m_1,...,m_N]$
$$
d_t=\sum\limits_{j=1}^Nz_jm_j.
$$
Based on $d_t$ and $s_t$, we compute a topic-aware state vector $o_t=d_t+s_t$ to integrate the latent topics, visual features and semantic information from the previously generated characters when predicting $y_t$.
\\
\\
Finally, the $t$-th character $y_t$ is predicted based on $o_t$, $\hat{v}_t$, $\hat{h}_t$ and $y_{t-1}$. Moreover, to enhance the topic consistency between poems and images, we encourage the decoder to generate characters that appear in the keywords extracted from the images by adding a bias probability to the characters in $K$. Specifically, we define a generic vocabulary $E_G$ which contains all possible characters, and a topic vocabulary $E_T$ which contains all characters in $K$, satisfying $E_T\subseteq E_G$. To predict $y_t$, we calculate the topic character probability $p_T(y_t)$ in addition to the generic character probability $p_G(y_t)$ by
\begin{align*}
p_G(y_t=w)&=g_G(o_t,\hat{v}_t,\hat{h}_t),\ \ \ \ \ \ w\in E_G,\\
p_T(y_t=w)&=\begin{cases}
g_T(o_t,\hat{v}_t,\hat{h}_t),& \mbox{$w\in E_T$,} \\
0,& \mbox{$w\in E_G\setminus E_T$},
\end{cases}
\end{align*}
where $g_T$ and $g_G$ represent functions corresponding to multilayer perceptrons followed by softmax, to compute $p_T$ and $p_G$ respectively.
$p_T$ and $p_G$ are summed to compute the character probability $p$, and the character with the maximum probability is picked as the next character:
\begin{align*}
p(y_t=w)&=\lambda p_T(y_t=w)+p_G(y_t=w),\\
y_t&=\arg\max_{w}p(y_t=w),w\in E_G,
\end{align*}
where $\lambda$ is a hyperparameter to balance the generic probability $p_G$ and the topic bias probability $p_T$.

\section{Experiments}
\subsection{Dataset}

In this paper, we are interested in the task of generating quatrains given images. To investigate the performance of our proposed framework on this task, a dataset of image-poem pairs needs to be constructed where images and poems are matched. For the convenience of data preparation, we focus on generating quatrains with 4 lines and 7 characters in each line. The framework proposed in this paper can be easily generalized to generate other types of poetry.

To construct the dataset of image-poem pairs, we collect 68,715 images from the Internet and use the poem dataset provided in~\cite{zhang2014chinese}, which contains 65,559 7-character quatrains.
Given the large numbers of images and poems, it is impractical to match them manually. Instead, we exploit the key concepts of the images and poems and match them automatically. Specifically, for each image, we obtain several keywords (e.g., water, tree) with the \textit{general-v1.3 model} provided by Clarifai; similarly we extract key concepts from each line of poems. 
Images and poem lines with common concepts can then be matched. In this way, a dataset with 2,311,359 samples is constructed, each of which consists of an image, the preceding poem lines and the next poem line. 
We randomly select 50,000 samples for validation, 1,000 for testing and the rest for training.


\begin{table*}[t]
\centering
\caption{Human evaluation of all models. Bold values indicate the best performance.}
\begin{tabular}{|c|c|c|c|c|c|c|} \hline
\textbf{Models} & \textbf{Poeticness} & \textbf{Fluency} & \textbf{Coherence} & \textbf{Meaning} & \textbf{Consistency} & \textbf{Average}\\ \hline\hline
SMT & 6.97 & 5.85 & 5.18 & 5.22 & 5.15 & 5.67\\
RNNPG-A & 7.27 & 6.62 & 5.95 & 5.91 & 5.09 & 6.17\\
RNNPG-H & 7.36 &  6.20 & 5.51 & 5.67 & 5.41 & 6.03\\
PPG-R & 6.47 & 5.27 & 4.91 & 4.82 & 4.96 & 5.29\\
PPG-H & 6.52 & 5.57 & 5.24 & 5.09 & 5.48 & 5.58 \\ \hline
MIPG (full) & \textbf{8.30} & \textbf{7.69} & \textbf{7.07} & \textbf{7.16} & \textbf{6.70} & \textbf{7.38}\\
MIPG (w/o keywords) & 7.21 & 6.78 & 6.13 & 6.15 & 4.10 & 6.08 \\
MIPG (w/o visual) & 7.05 & 4.81 & 4.76 & 5.21 & 3.98 & 5.16\\ \hline
\end{tabular}
\label{human_evaluation}
\end{table*}
\subsection{Training Details}
We use 6,000 most frequently used characters as the vocabulary $E_G$. The number of recurrent hidden units is set to 512 for both encoder and decoder. The dimensions of the input and output memory in the topic memory network are also both set to 512. The hyperparameter $\lambda$ to balance $P_T$ and $P_G$ is set to 0.5 which is tuned on the validation set from $\lambda=0.0,0.1,...,1.0$.
All parameters are randomly initialized from a uniform distribution with support from [-0.08, 0.08]. The model is trained with the AdaDelta algorithm~\cite{zeiler2012adadelta} with the batch size set to 128, and the final model is selected according to the cross entropy loss on the validation set. During training, we invert each line to be generated following \cite{yi2016generating} to make it easier for the model to generate poems obeying rhythm rules. For the visual feature extractor, we choose a pre-trained VGG-19~\cite{simonyan2014very} model and use the output of  the \textit{conv5\_4} layer, which includes 196 vectors of 512 dimensions, as the local visual features of an image. For the majority of images, 10-20 keywords are extracted.


\subsection{Evaluation Metrics}

For the general task of text generation in natural language processing, there exist various metrics for evaluation including BLEU and ROUGE. However, it has been shown that the overlap-based automatic evaluation metrics have little correlation with human evaluation~\cite{liu2016not}. Thus, for automatic evaluation, we only calculate the recall rate of
the key concepts in an image which are described in the generated poem to evaluate whether our model can generate consistent poems given images. Considering the uniqueness of the poem generation task in terms of text structure and literary creation, we evaluate the quality of the generated poems with a human study.


Following~\cite{zhewang2016chinese}, we use the metrics listed below to evaluate the quality of a generated poem:
\begin{itemize}
\item \textbf{Poeticness}. Does the poem follow the rhyme and tone regulations?
\item \textbf{Fluency}. Does the poem read smoothly and fluently?
\item \textbf{Coherence}. Is the poem coherent across lines?
\item \textbf{Meaning}. Does the poem have a reasonable meaning and artistic conception?
\end{itemize}

In addition to these metrics, for our task of generating Chinese classical poems given images, we need to evaluate how well the generated poem conveys the input image. Here we introduce a metric \textbf{Consistency} to measure whether the topics of the generated poem and the given image match.
\subsection{Model Variants}
In addition to the proposed framework of MIPG, we evaluate two variants of the model to examine the influence of the visual and semantic topic information on the quality of the poems generated:
\begin{itemize}
\item \textbf{MIPG (full)}. The proposed model, which integrates direct visual information and semantic topic information.
\item \textbf{MIPG (w/o keywords)}. Based on MIPG, the semantic topic information is removed by setting the input and output memory vectors of keywords to $\vec{0}$, such that the model only leverages visual information in poetry generation.
\item \textbf{MIPG (w/o visual)}. Based on MIPG, the visual information is removed by setting the visual feature vectors of image to $\vec{0}$, such that the model only leverages semantic topic information in poetry generation.
\end{itemize}

\subsection{Baselines}
As far as we know, there is no prior work on generating Chinese classical poetry from images. Therefore, for the baselines we implement several previously proposed keyword-based methods listed below:

\textbf{SMT}. A statistical machine translation model~\cite{he2012generating}, which translates the preceding lines to generate the next line, and the first line is generated from input keywords with a template-based method.

\textbf{RNNPG}. A poetry generation model based on recurrent neural networks~\cite{zhang2014chinese}, where the first line is generated with a template-based method taking keywords as input and the other three lines are generated sequentially. In our implementation, two schemes are used to select the keywords for the first line, including using all the keywords (RNNPG-A) and using the important keywords selected by human (RNNPG-H) respectively. Specifically, for RNNPG-A, we use all the keywords extracted from an image (10-20 keywords), while for RNNPG-H, we invite 3 volunteers to vote for the most important keywords in the image (3-6 keywords).

\textbf{PPG}. An attention-based encoder-decoder framework with a sub-topic keyword assigned to each line~\cite{zhewang2016chinese}. Since PPG requires that the number of keywords to be equal to the number of lines (4 for quatrains), we consider two schemes to select 4 keywords from the keyword set extracted from the image, including random selection (PPG-R) and human selection (PPG-H). Specifically, for PPG-R, we randomly select 4 keywords from the keyword set and randomly sort them. For PPG-H, we invite 3 volunteers to vote for the 4 most important keywords in the image and sort them in the order of relevance.

\begin{table}[t]
\centering
\caption{Recall rate of the key concepts in an image described in the poems generated by all the models.}
\begin{tabular}{|c|c|c|c|} \hline
\textrm{Models} & Recall & \textrm{Models} & \textrm{Recall}\\ \hline \hline
RNNPG-A & 12.85\%  & PPG-R & 33.5\% \\ \hline
RNNPG-H & 11.82\% & PPG-H & 33.7\% \\ \hline
SMT & 19.79\% & MIPG & \textbf{58.8\%} \\ \hline
\end{tabular}
\label{consistency_recall}
\end{table}

\begin{table*}\small
\centering
\begin{tabular}{|c|c|} \hline
 & \\
\includegraphics[scale = 2.2]{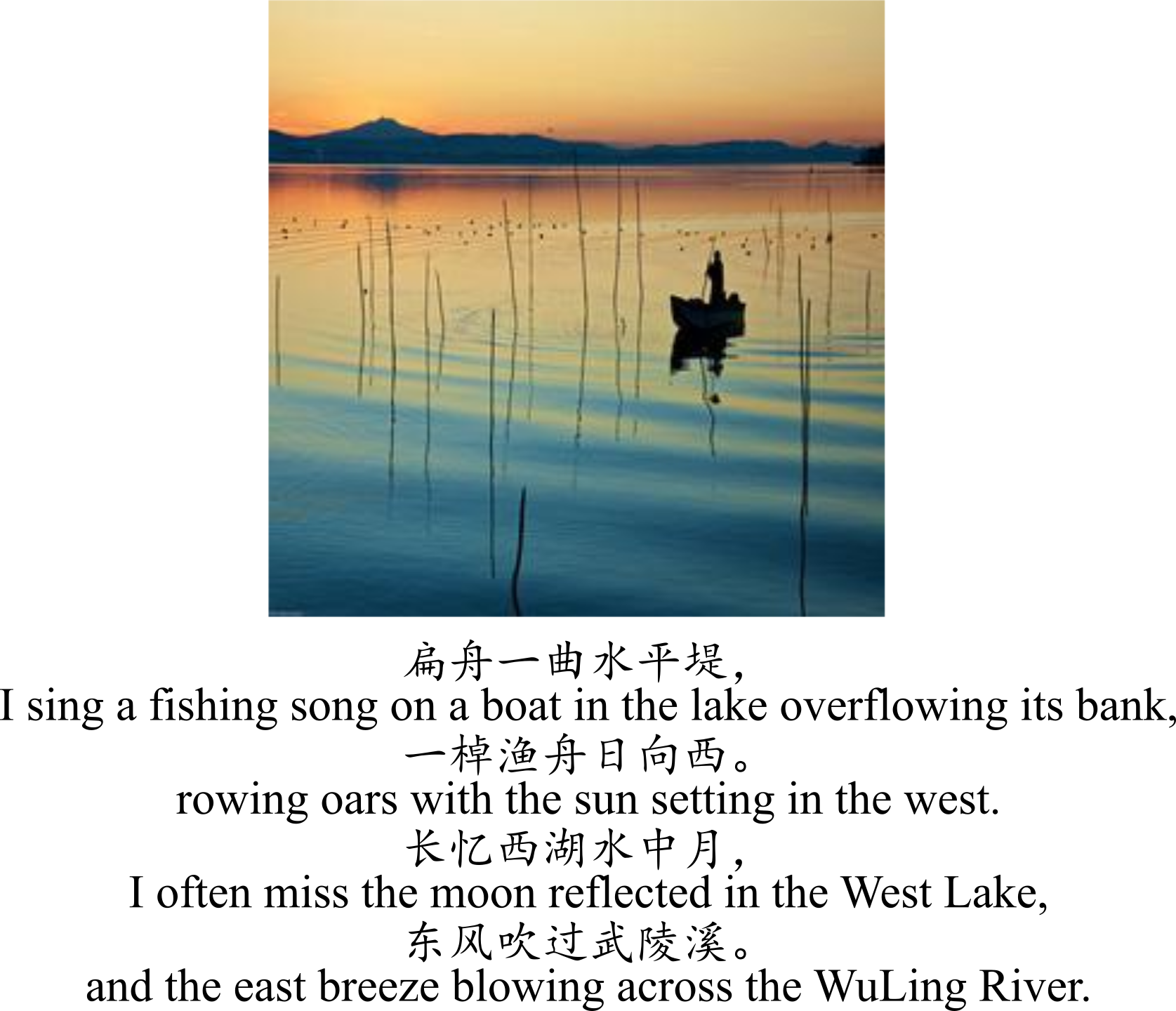} & \includegraphics[scale = 2.2]{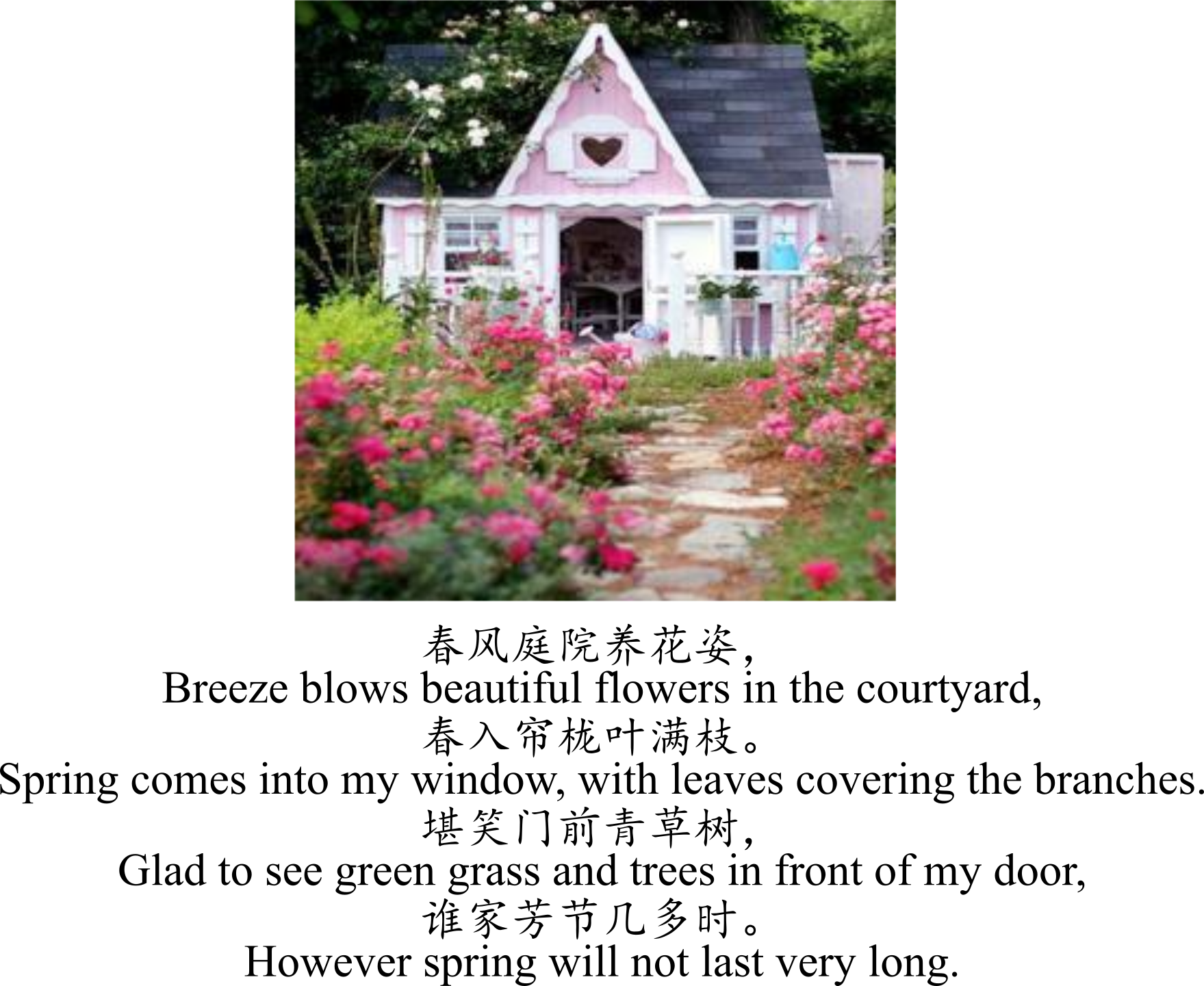} \\
\hline
\end{tabular}
\caption{Two sample poems generated from the corresponding image by MIPG.}
\label{example}
\end{table*}

\subsection{Human Evaluation}
We invite eighteen volunteers, who are knowledgeable in classical Chinese poetry from reading to writing, to evaluate the results of various methods. 45 images are randomly sampled as our testing set. Volunteers rate every generated poem with a score from 1 to 10 from 5 aspects: Poeticness, Fluency, Coherence, Meaning and Consistency. Table~\ref{human_evaluation} summarizes the results.
\subsubsection{Overall Performance.}
The results in Table~\ref{human_evaluation} indicate that the proposed model MIPG outperforms the baselines with all metrics. It is worth mentioning that in terms of ``Consistency'' which measures how well the generated poem can describe the given image, MIPG achieves the best performance, demonstrating the effectiveness of the proposed model at capturing the visual information and semantic topic information in the generated poems. 
From the comparison of RNNPG-A and RNNPG-H, one can notice that, by picking important keywords manually, 
the poems generated by RNNPG-H are more consistent with images, which implies the importance of keyword selection in poetry generation. Similarly, with important keywords selected manually, PPG-H outperforms PPG-R from all aspects, especially in ``Coherence'' and ``Consistency''. As a comparison, in the proposed MIPG framework, the topic memory network makes it possible to dynamically determine a topic while generating each character, in the meantime, the encoder-decoder framework ensures the generation of fluent poems following the regulations. As a whole, the visual information and the topic memory network work together to generate poems consistent with given images.

\subsubsection{Analysis of Model Variants.}

The results in the bottom rows of Table~\ref{human_evaluation} correspond to the model variants including MIPG (full), MIPG (w/o keywords) and MIPG (w/o visual). One can observe that ignoring semantic keywords in MIPG (w/o keywords) or visual information in MIPG (w/o visual) degrades the performance of the proposed model significantly, especially in terms of ``Consistency''. This provides clear evidence justifying that the visual information and semantic topic information work together to generate poems consistent with images.

\subsection{Automatic Evaluation of Image-Poem Consistency}

Different from the keyword-based poetry generation models, for the task of poetry generation from images, the image-poem consistency is a new and very important metric while evaluating the quality of the generated poems. Therefore, we conduct an automatic evaluation in terms of ``Consistency'' in addition to human evaluation, which is achieved by computing the recall rate of the key concepts in an image that are described in the generated poem.

The results are shown in Table~\ref{consistency_recall}, where one can observe that the proposed MIPG framework outperforms all the other baselines with a large margin, which indicates that the poems generated by MIPG can better describe the given images.
One should also notice the difference between the subjective and quantitative evaluation in ``Consistency'' from Table~\ref{human_evaluation} and Table~\ref{consistency_recall}. For instance, PPG-R and PPG-H achieve almost equal recall rates of keywords, but the ``Consistency'' score of PPG-R is significantly lower than PPG-H in Table~\ref{human_evaluation}. This is reasonable considering that both PPG-R and PPG-H select 4 keywords extracted from the given image to generate poems, which yield equal recall rates. In the meantime, the keywords selected by human describe the image better than randomly picked keywords, therefore PPG-H can generate poems more consistent with images than PPG-R from the human perspective. In addition, the reason that the keyword recall rates of RNNPG-A and RNNPG-H are relatively low is probably due to the fact that RNNPG often generates the first poem line which is semantically relevant with given keywords while not containing any of them. As a consequence, RNNPG-A and RNNPG-H may generate poems with low keyword recall rates but high ``Consistency'' scores.

\subsection{Examples}

To further illustrate the quality of the poems generated by the proposed MIPG framework, we include two examples of the poems generated with the corresponding images in Table~\ref{example} . As shown in these examples, the poems generated by MIPG can capture the visual information and semantic concepts in the given images. More importantly, the poems nicely describe the the images in a poetic manner, while following the strict regulations of classical Chinese poetry.

\section{Conclusion}
In this paper, we propose a memory based neural network model for classical Chinese poetry generation from images (MIPG) 
where visual features as well as semantic topics of images are exploited when generating poems. Given an image, semantic keywords are extracted as the skeleton of a poem, where a topic memory network is proposed that can take as many keywords as possible and dynamically select the most relevant ones to use during poem generation. On this basis, visual features are integrated to embody the information missing in the keywords. Numerical and human evaluation regarding the quality of the poems from different perspectives justifies that the proposed model can generate poems that describe the given images accurately in a poetic manner, while following the strict regulations of classical Chinese poetry.
\section{Acknowledgements}
This research was supported by the National Natural Science Foundation of China (No. 61375060, No. 61673364, No. 61727809 and No. 61325010), and the Fundamental Research Funds for the Central Universities (WK2150110008). We also gratefully acknowledge the support of NVIDIA Corporation with the donation of the Titan X GPU used for this work.

\bibliographystyle{aaai}
\bibliography{aaai}
\end{document}